\documentclass{article}

\usepackage[preprint]{neurips_2025}

\usepackage[utf8]{inputenc} %
\usepackage[T1]{fontenc}    %
\usepackage{hyperref}       %
\usepackage{url}            %
\usepackage{booktabs}       %
\usepackage{amsfonts}       %
\usepackage{nicefrac}       %
\usepackage{microtype}      %
\usepackage{xcolor}         %

\usepackage{graphicx}  %
\usepackage{subcaption} %
\usepackage{lipsum}     %
\usepackage{xspace}

\newcommand{\name}{SpecReason\xspace}
\newcommand{\squishlist}{
   \begin{list}{$\bullet$}
    { \setlength{\itemsep}{0pt}      \setlength{\parsep}{3pt}
      \setlength{\topsep}{3pt}       \setlength{\partopsep}{0pt}
      \setlength{\leftmargin}{1.0em} \setlength{\labelwidth}{1em}
      \setlength{\labelsep}{0.5em} } }
\newcommand{\squishend}{
    \end{list}  }

\title{SpecReason: Fast and Accurate Inference-Time Compute via Speculative Reasoning}

\author{
  \textbf{Rui Pan}$^{\mathsection}$
  \quad
  \textbf{Yinwei Dai}$^{\mathsection}$
  \quad
  \textbf{Zhihao Zhang}$^{\dag}$
  \quad
  \textbf{Gabriele Oliaro}$^{\dag}$
  \\
  \textbf{Zhihao Jia}$^{\dag}$
  \quad
  \textbf{Ravi Netravali}$^{\mathsection}$
  \\
  ${}^\mathsection$Princeton University \quad 
  ${}^\dag$Carnegie Mellon University\\
    \texttt{\{ruipan,yinweid\}@princeton.edu}, \texttt{\{zhihaoz3,goliaro\}@cs.cmu.edu},\\
    \texttt{zhihao@cmu.edu}, \texttt{rnetravali@cs.princeton.edu}
}

\begin{document}

\maketitle

\begin{abstract}

Recent advances in inference-time compute have significantly improved performance on complex tasks by generating long chains of thought (CoTs) using Large Reasoning Models (LRMs). However, this improved accuracy comes at the cost of high inference latency due to the length of generated reasoning sequences and the autoregressive nature of decoding. Our key insight in tackling these overheads is that LRM inference, and the reasoning that it embeds, is highly tolerant of approximations: complex tasks are typically broken down into simpler steps, each of which brings utility based on the semantic insight it provides for downstream steps rather than the exact tokens it generates. Accordingly, we introduce \name{}, a system that automatically accelerates LRM inference by using a lightweight model to (speculatively) carry out simpler intermediate reasoning steps and reserving the costly base model only to assess (and potentially correct) the speculated outputs. Importantly, \name{}'s focus on exploiting the semantic flexibility of thinking tokens in preserving final-answer accuracy is complementary to prior speculation techniques, most notably speculative decoding, which demands token-level equivalence at each step. Across a variety of reasoning benchmarks, \name{} achieves $1.4-3.0\times$ speedup over vanilla LRM inference while improving accuracy by $0.4-9.0\%$. Compared to speculative decoding without \name{}, their combination yields an additional $8.8-58.0\%$ latency reduction. We open-source SpecReason at \url{https://github.com/ruipeterpan/specreason}.

\end{abstract}

\section{Introduction}

Inference-time compute has unlocked a new axis for scaling AI capabilities. Recent advancements in Large Reasoning Models (LRMs) such as OpenAI o1/o3~\citep{openai_o1,openai_o3} and DeepSeek R1~\citep{deepseek_r1} have demonstrated state-of-the-art performance across a wide range of complex tasks. 
Although these LRMs share the architectural backbones as traditional large language models (LLMs), their inference behavior differs significantly: LRMs first ``think'' by generating internal {\em thinking} tokens---tokens that decompose a task into a sequence of composable reasoning steps via a long chain-of-thought (CoT)~\citep{cot} before producing the final tokens that summarize the reasoning process.

Despite their promise, LRMs incur substantial inference latency due to the length of the reasoning sequences they generate. This challenge is primarily driven by the autoregressive nature of LLMs, where decoding time scales linearly with sequence length. As a result, final output generation can routinely take minutes, if not hours, to answer a single query; such delays far exceed those of typical LLMs and are prohibitively slow for many interactive applications, ultimately degrading user experience~\citep{certaindex}.

\begin{figure}[t]
    \centering
    \includegraphics[width=0.99\textwidth]{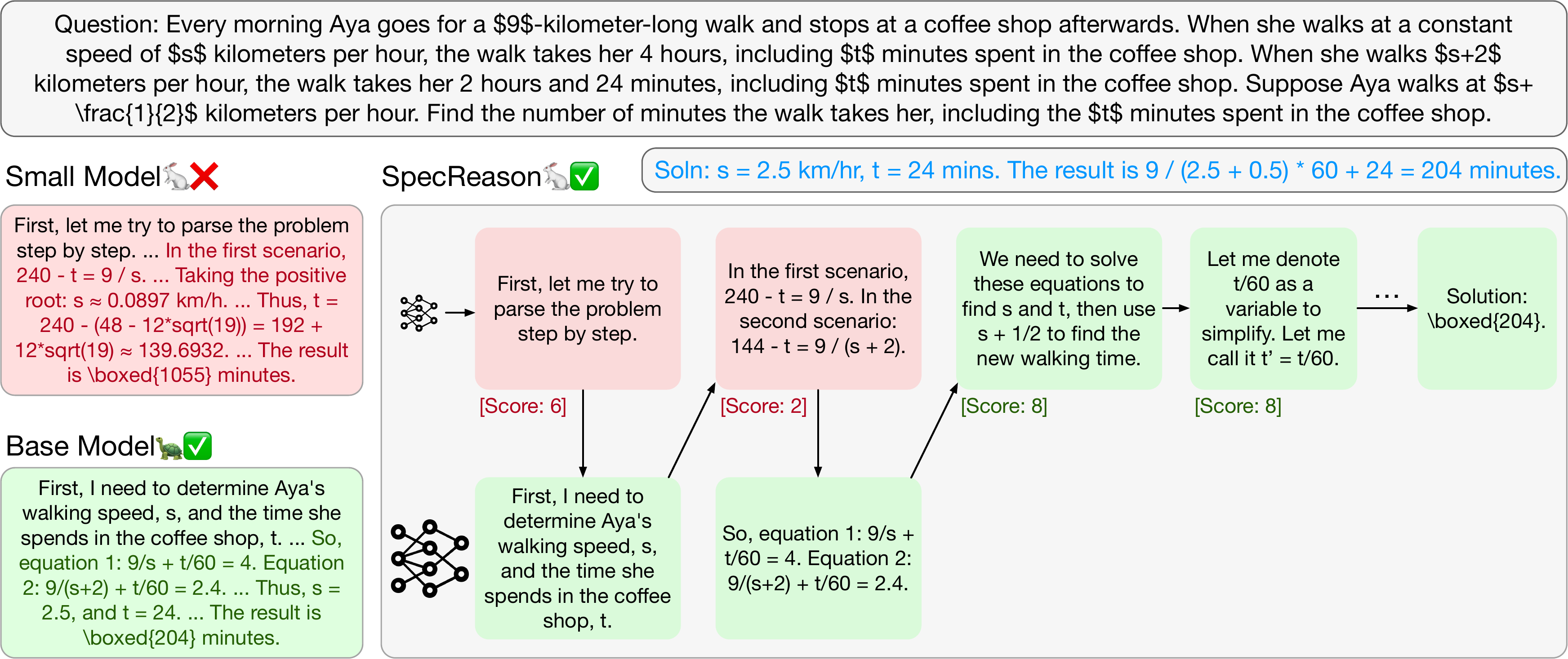}
    \caption{\name{} leverages a smaller reasoning model to speculate individual reasoning steps, deferring to the base model only for assessment (and optionally as a fallback), enabling faster yet accurate reasoning. For illustration, we show a math question as an example; our evaluation includes more general reasoning workloads.}
    \label{fig:toy_example}
\end{figure}

Our approach to tackling reasoning delays---{\bf without compromising accuracy}---is rooted in two fundamental properties of LRMs: (1) LRMs tackle difficult tasks by generating long CoTs that decompose them into many simpler, sequential steps. For example, in mathematical problem solving, a few key reasoning steps require complex long-term planning and have a major influence on downstream reasoning, while most subsequent steps simply execute the plan through straightforward calculations or case analyses (Fig.~\ref{fig:toy_example});
(2) The utility of an individual reasoning step hinges less on the exact wording of the thinking tokens but more on the {\em semantic insight} it provides.
That is,  as long as a step contributes meaningfully to advancing the CoT, it remains effective—even if phrased imprecisely or differently (Fig.~\ref{fig:equiv_spectrum}).
Moreover, LRMs possess self-reflection capabilities that enable them to revise or correct occasional missteps from earlier steps. 

\textbf{Taken together, these properties make the decoding of thinking tokens---the dominant source of inference latency in LRMs---inherently more \emph{approximation tolerant} than typical LLM decoding}. A large fraction of intermediate reasoning steps can be effectively handled by lightweight reasoning models, which both align with the nature of these steps and can tolerate minor inaccuracies. As shown in Fig.~\ref{fig:main_results}, this opens the door to significantly faster inference without sacrificing output quality.

Building on these insights, we propose \textbf{\name{}}, a system for accelerating LRM inference by selectively offloading easier intermediate steps to be {\em speculated} by a smaller model without compromising final output accuracy. 
\name{} employs a lightweight reasoning model to generate individual reasoning steps, while reserving the slower but more capable base model to efficiently verify these speculated steps (\S\ref{sec:spec_reason}) and guide the reasoning process along the correct trajectory (Fig.~\ref{fig:toy_example}). 
Consistent with prior findings~\citep{prmbench}, we observe that base models can be prompted to act as critic models---assessing the utility of intermediate steps and accepting or rejecting them as needed (Fig.~\ref{fig:prm_microbenchmark}). %

\textbf{Speculative reasoning vs. speculative decoding.} While \name{} is conceptually related to speculative decoding~\citep{spec_decoding}, which accelerates LLM inference by using a smaller draft model to predict future tokens, there are key distinctions between the two. Most notably, speculative decoding is an \emph{exact} optimization: it relies on {\em token}-level equivalence between the small and base models, i.e., focusing on typical LLM serving where all generated tokens are part of the final model output being assessed. In contrast, \name{} explicitly leverages the {\em approximation tolerance} inherent in reasoning: it targets {\em thinking tokens}---intermediate steps in the reasoning process---where semantic alignment, rather than token-level equivalence, is sufficient.
This relaxation enables substantial latency savings during LRM inference, as semantically similar intermediate steps (Fig.~\ref{fig:equiv_spectrum}) are often adequate to preserve end-task accuracy (Fig.~\ref{fig:main_results}).
In many cases, \name{} even \emph{improves} final accuracy over the base model by generating fewer unnecessary tokens (Fig.~\ref{fig:acc_intuition}).
To further address the high inference cost of LRMs, \name{} also exposes a user-configurable knob that allows trading off accuracy for latency by adjusting the tolerance level for speculative approximations.
Finally and most importantly, because speculative reasoning and speculative decoding operate at different levels, we show that they are \emph{complementary} techniques (\S\ref{sec:hierarchical_speculation}), and when combined in a hierarchical speculation framework, achieve even greater reductions in inference latency.

We evaluate \name{} across a wide range of reasoning workloads spanning tasks of varying complexity~\citep{aime,math500,gpqa}. Overall, \name{} reduces end-to-end inference latency by $1.4-3.0\times$ compared to vanilla LRM inference while improving accuracy by $0.4-9.0\%$. Moreover, \name{} can be {\em combined} with speculative decoding to provide an additional $8.8-58.0\%$ improvement over speculative decoding alone.

\section{Background}

\textbf{Inference-time scaling.} LRMs introduce a structured problem-solving approach that breaks down complex problems into multiple simpler reasoning steps, commonly referred to as a long chain of thought (CoT)~\citep{cot}. This enables the model to generate intermediate reasoning steps before progressing further, reflect, and backtrack to correct errors if needed. LRMs that output long CoTs have been a popular approach to scale inference-time compute~\citep{deepseek_r1,openai_o1,openai_o3}, and there also exist other schemes like Tree of Thoughts~\citep{tot}, process-reward-model-guided tree search~\citep{prm,rstar,rstar_math}, and repeated sampling for scaling inference-time compute~\citep{large_language_monkeys}.

\textbf{Speculative decoding.} Speculation has long been a classic concept in the literature of computer architecture~\citep{burton1985speculative}. Due to the memory-bound nature of LLM decoding, recent work has also leveraged the technique of speculation to accelerate the decoding phase~\citep{stern2018blockwise, spec_decoding, decoding_specdecoding} of LLM inference. The speculative decoding process alternates between speculation and verification steps to ensure correctness while achieving speed-ups. The speculation phase usually consists of either a standalone draft model~\citep{spec_decoding, specinfer}, a trainable module on top of the base model~\citep{medusa, eagle}, a tree-based token cache~\citep{suffix_decoding, token_recycling, lookahead_alipay}, an n-gram lookup table~\citep{lookahead}, or a retrieval-based data store~\citep{rest} to make efficient but less accurate speculations. The verification process, on the other hand, is a base model chunked-prefill over the speculation results, which usually consists of either a single sequence of tokens as in~\citet{spec_decoding} or tree-like structures to further boost the accuracy of speculation~\citep{specinfer, medusa, eagle, sequoia}. The verification process then accepts the longest matched sequences on the token level from the speculation results and repeats the process. As a result, the speculation length is usually conservative to maintain an optimal trade-off between the speculation overhead and accuracy. 

\textbf{Existing approaches for reducing latency.} Sky-T1-Flash~\cite{sky_t1_flash} reduces unnecessary thinking tokens by fine-tuning models to curb overthinking, thereby reducing the length of reasoning chains and, consequently, latency. Dynasor-CoT~\cite{certaindex,dynasor_cot} takes a different approach by probing intermediate model confidence and terminating the reasoning process early when the model exhibits sufficient confidence in its current output.

\section{Motivation}

\begin{figure}[t]
    \centering
    \includegraphics[width=0.9\textwidth]{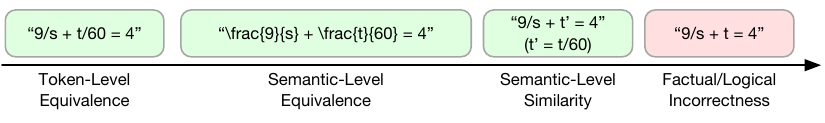}
    \caption{The spectrum of approximations of one example reasoning step (equation 1 in Fig.~\ref{fig:toy_example}). \name{} can control the exactness of reasoning approximations by adjusting its acceptance threshold to navigate through the accuracy-latency tradeoff space (\S\ref{sec:acc_lat_tradeoff}).}
    \label{fig:equiv_spectrum}
\end{figure}

In this work, we show that reasoning workloads executed by LRMs exhibit unique opportunities for latency reduction due to their inherent tolerance to approximation--- setting them apart from traditional generation tasks in LLMs. We illustrate these properties using a representative example from the AIME dataset, selected for its clarity and ease of exposition.

\textbf{Intermediate steps are easier than end-to-end reasoning.} A key observation in LRM behavior is that reasoning difficulty is not uniform across the steps in a long chain-of-thought (CoT). As shown in Fig.~\ref{fig:toy_example}, while the overall task might be too challenging for a small model to solve end-to-end, only a few critical steps---such as problem analysis, decomposition through formulations or case analyses, and high-level planning---are critical to the overall reasoning progress. In contrast, many other steps are significantly easier.

This behavior is intentional by design: LRMs are often trained with reinforcement learning to generate CoTs that decompose complex problems into sequences of simpler, more tractable reasoning steps. These intermediate steps often include routine reasoning such as arithmetic calculations, case enumeration, or basic logical deductions---operators that are much easier to decode than synthesizing a full solution directly. This heterogeneity in step difficulty and importance creates an opportunity for lightweight models to handle a substantial portion of the reasoning process both efficiently and accurately.

\textbf{Reasoning progress depends on insights, not exact tokens.} Another key takeaway from our work is that the utility of a reasoning step lies in the semantic contribution it makes to the overall reasoning process, rather than the precise tokens it uses. Unlike tasks like translation in traditional LLM inference, where fidelity to exact combinations of tokens matters more, reasoning CoTs within LRM's thinking tokens care more about the information that advances the reasoning chain. 
As illustrated in Fig.~\ref{fig:equiv_spectrum}, a spectrum of valid phrasings often exists for a given step: semantically equivalent or similar expressions can convey the same insight and lead to the same downstream reasoning trajectory. This semantic flexibility is a key enabler for approximation-tolerant inference.

\textbf{Occasional mistakes can be corrected via self-reflection.} LRMs exhibit strong self-reflection capabilities, enabling them to recover from earlier reasoning errors. 
Even when an earlier step contains a factual or logical mistake, the model often revises its trajectory in subsequent steps, marked by tokens like ``Wait'' or ``Hmm''. Moreover, unlike LLM inference where {\em all} output tokens contribute to the final answer, in LRM inference, only the tokens generated {\em after} the thinking tokens determine the final outcome. Therefore, LRM inference can tolerate occasional mistakes during the reasoning phase, as the model can often identify and correct these mistakes during self-reflection. This inherent fault tolerance further underscores the viability and effectiveness of approximation-based acceleration.

In summary, compared to traditional LLM inference, LRM inference is inherently more tolerant of approximations that do not require token-level equivalence as long as the overall reasoning trajectory is preserved. This property is not limited to a single, linear CoT; rather, it extends naturally to more general inference-time compute scaling paradigms such as tree-based search strategies and other structured reasoning approaches.

\section{Method}

\subsection{Speculative Reasoning}
\label{sec:spec_reason}

Due to its reliance on autoregressive decoding, LRM inference incurs significantly higher latency than typical LLMs---often to the point of being prohibitively slow for interactive applications and degrading user experience~\citep{dynasor_cot}. Existing approaches for latency reduction include using a distilled version of the base model~\citep{deepseek_r1}, limiting the number of thinking tokens via a predefined {\em token budget}, or disabling the reasoning process altogether by omitting the thinking tokens (\texttt{<think>} and \texttt{</think>}) during generation~\citep{qwen3}. However, these approaches impose a harshly trade-off between accuracy for latency: they either limit the model's capacity to reason or apply a lower-quality model uniformly across all reasoning steps.
In contrast, \name{} takes a more fine-grained and adaptive approach. Instead of explicitly restricting output length, it selectively offloads only the easier reasoning steps to a lightweight model, preserving overall reasoning quality while substantially reducing inference latency.

The approximation-tolerant nature of LRM reasoning enables a new form of speculative execution: tentatively carrying out reasoning steps using a lightweight model, assessing their utility with a stronger base model, and selectively accepting them. \name{} leverages this flexibility to reduce decoding latency while preserving output quality.
To achieve this goal, \name{} offloads easier or less critical reasoning steps---defined as semantically self-contained units such as complete sentences or logical steps---to a smaller, faster {\em speculator} model. Each step is decoded in two stages: (1) the lightweight speculator proposes the next reasoning step based on the current context, and (2) the base model evaluates the proposed step for semantic utility. If the step is accepted, \name{} proceeds to the next step; otherwise, \name{} falls back to the base model to regenerate the step. While our implementation uses a simple static-threshold mechanism for verification, the framework supports richer, customizable decision strategies. We outline key design principles below.

\textbf{Navigating the Pareto frontier of the latency-accuracy tradeoff.} \name{} expands the Pareto frontier of the latency-accuracy tradeoff by exposing fine-grained control knobs to navigate through this space. The key knob \name{} employs is the acceptance threshold: after each speculated reasoning step, the base model is prompted to generate a single-token utility score (e.g., an integer from 0 to 9) indicating the quality of the step. If the utility score is above a static acceptance threshold (e.g., score $\geq 7$), the speculated reasoning step is accepted; otherwise, it is discarded and regenerated by the base model.

Adjusting this threshold allows users to control the {\em strictness} of speculation (Fig.~\ref{fig:lat_acc_tradeoff}): a higher threshold requires speculated steps to be closer to token-level equivalence on the equivalence spectrum (Fig.~\ref{fig:equiv_spectrum}), improving accuracy but reducing the acceptance rate and thereby increasing latency. Conversely, a lower threshold increases speculation efficiency at the cost of potential accuracy degradation.

An additional knob involves forcing the first $n$ reasoning steps to be decoded by the base model. Since LRMs often use the initial steps to analyze the problem and formulate a high-level plan, assigning these initial steps to the base model can steer the overall reasoning trajectory toward higher quality.
We show in Fig.~\ref{fig:earlyforce_knob} that this knob also allows \name{} to manage latency-accuracy tradeoff, though with less impact than the acceptance threshold knob.

While our current implementation uses a simple, discrete threshold-based scoring scheme---offering only a coarse-grained configuration space---it establishes a lower bound on verification quality. 
Future work can explore more sophisticated strategies, such as logprob-based confidence estimates or dynamic thresholds, to enable finer-grained tradeoffs without incurring additional runtime cost, and may further improve overall performance.

\textbf{Efficient verification.} Because each step requires verification by the base model, it's crucial to keep verification overhead low to avoid compounding latency. 
Instead of autoregressively decoding or reranking multiple candidate steps, \name{} evaluates each speculated step in a single \textit{prefill-only} pass of the base model. The verification prompt is templated to reuse most of the CoT prefix, so each verification requires prefilling only $\sim$70 new tokens. Since short-prefill forward passes are memory-bound, the overhead is comparable to decoding just 1–2 tokens, making verification highly efficient in practice.

\textbf{Implementation details.} Since the small model is lightweight, we colocate both the small and base models on the same GPU. The memory reserved for Key-Value caches~\citep{vllm} is statically partitioned between the two models. They do not share any internal model states--only the token IDs of the generated reasoning steps are managed and shared by \name{}. If a speculative step is rejected, the corresponding KV cache entries are discarded. 

Inference is performed sequentially: the small and base models take turns, avoiding kernel-level interference. In future work, we plan to explore pipelining to overlap the small model’s decoding with the base model’s inference. While this may introduce mild resource contention, it could further reduce end-to-end latency.

\subsection{Hierarchical Speculation across Semantic Similarity and Token Equivalence}
\label{sec:hierarchical_speculation}

At a high level, \name{}'s speculative reasoning resembles the philosophy behind traditional speculative decoding, but differs in two important ways. First, speculative decoding guarantees token-level equivalence between draft and verified outputs, making it a form of exact acceleration. In contrast, \name{} targets semantic-level similarity, accepting steps that carry the same insight even if phrased differently, and exposes knobs to control the exactness of reasoning approximations. Second, speculative decoding is typically applied to output generation tasks (e.g., text continuation or translation), where the fidelity of each token matters. \name{}, on the other hand, is designed specifically for internal thinking tokens in reasoning tasks, where intermediate steps are approximate and interchangeable as long as they preserve the logical progression of thought.

Further, because \name{} and speculative decoding operate at different levels (semantic-level similarity vs. token-level equivalence), these two approaches are complementary and can be combined into a unified, hierarchical system -- \name{}+Decode first applies step-level speculative reasoning to draft and verify reasoning steps. If a step is rejected and regenerated by the base model, standard token-level speculative decoding can be applied during the base model regeneration to further accelerate decoding. %

\section{Evaluation}

\begin{figure}[t]
    \centering
    \begin{subfigure}[t]{0.5\textwidth}
        \centering
        \includegraphics[width=\linewidth]{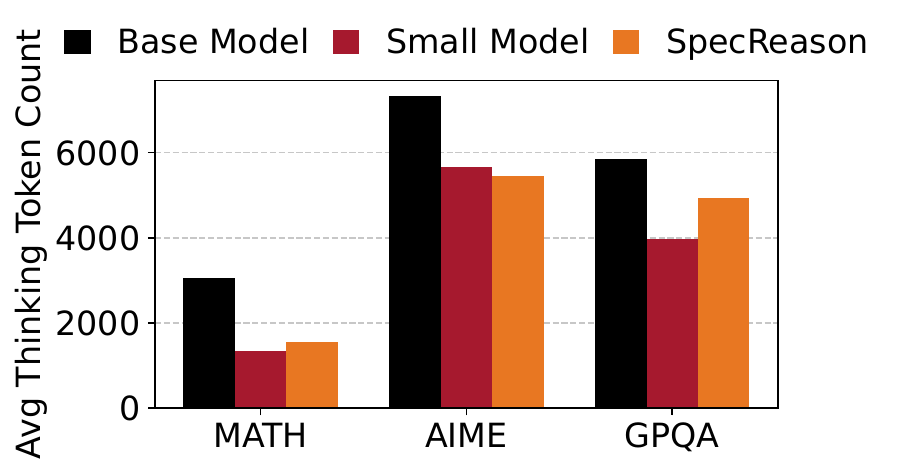}
        \caption{Output length comparison. \name{} reduces the token consumption needed to answer queries by adopting speculated steps from small models that are less verbose.}
        \label{fig:output_len_comparison}
    \end{subfigure}
    \hfill
    \begin{subfigure}[t]{0.48\textwidth}
        \centering
        \includegraphics[width=\linewidth]{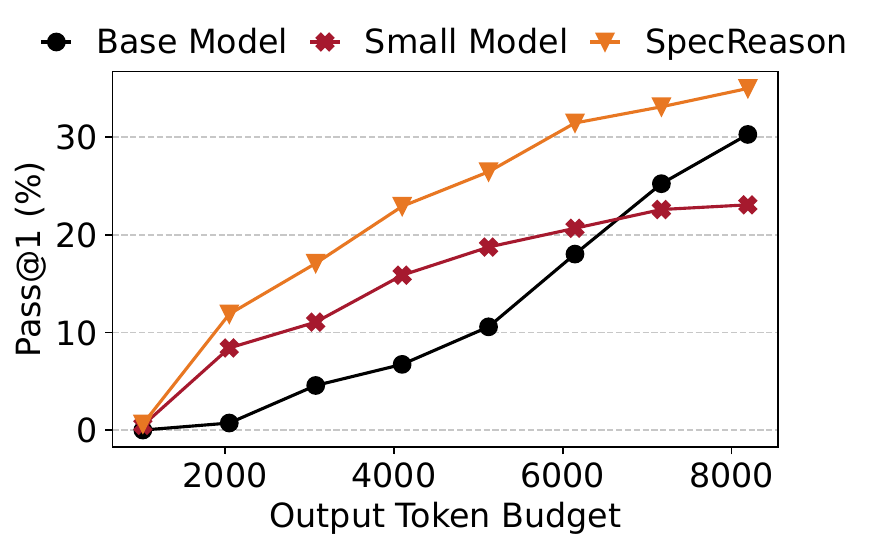}
        \caption{[AIME] Accuracy gap under different token budgets.}
        \label{fig:acc_by_seq_len}
    \end{subfigure}
    \caption{[QwQ-32B + Zyphra-1.5B] Intuition behind \name{}'s accuracy improvement. See Fig.~\ref{fig:acc_insight_len_all} in \S\ref{sec:appendix} for the full set of results.}
    \label{fig:acc_intuition}
\end{figure}

The overview of our evaluation results include:

\squishlist
    \item \textbf{Reducing end-to-end latency.} Because many intermediate steps are easier than end-to-end reasoning, many (up to 80\%) of the speculated steps are accepted. \name{} achieves a $1.4-3.0\times$ speedup over vanilla LRM inference. Additionally, when combined with speculative decoding, \name{} further reduces latency by $8.8-58.0\%$ over speculative decoding alone, highlighting the complementary nature of these optimizations.
    \item \textbf{Improving token-budget-aware accuracy.} Beyond latency reduction, \name{} also improves accuracy over the base model by $0.4-9.0\%$ under the same token budget. We empirically find that small, lightweight models typically have shorter output sequence lengths -- meaning, they need fewer thinking tokens before deriving an answer. Thus, by accepting many small model's speculated reasoning steps, \name{} reduces the token consumption compared to the base model's vanilla inference. When the token budget is low -- a common setup to curb inference cost and latency -- \name{} helps improve accuracy as the base model would need more tokens to get to an answer (Fig.~\ref{fig:acc_intuition}).
\squishend

\subsection{Setup}

\textbf{Models.} In our main results, we use two base models: QwQ-32B~\citep{qwq_32b} and Skywork-OR1-Preview-32B~\citep{skywork_32b}. We also use two different small models for speculation: DeepSeek-R1-1.5B~\citep{deepseek_r1} and Zyphra's ZR1-1.5B~\citep{zyphra_1.5b} -- both of which are based on Qwen-2.5~\citep{qwen2.5} and embed the capability of reasoning with long CoTs -- and evaluate all four different model combinations. We evaluate an additional base model with a different size and architecture, R1-70B~\citep{deepseek_r1}, a distilled version of DeepSeek-R1 onto Llama3.3-70B~\citep{llama3}, in~\S\ref{sec:70b_microbenchmark}.

\textbf{Datasets.} We evaluate \name{} on three diverse reasoning benchmarks: AIME~\citep{aime} for high-school competition-level mathematical problems, MATH500~\citep{math500} for high-school competition-level mathematical problems sampled from AMC 10, AMC 12, and AIME, and GPQA Diamond~\citep{gpqa} for graduate-level questions in general domains like biology, physics, and chemistry. The accuracy metric we evaluate on is pass@1. Similar to prior work~\citep{deepseek_r1}, we set k=16 when calculating pass@1 -- i.e., we generate 16 responses with temperature=0.6 for every query and calculate the average accuracy -- and set the token budget to be 8192 tokens to ensure an apples-to-apples comparison between baselines.  %

\textbf{Baselines.} We run vanilla inference using the small and base models as the latency and accuracy baseline, respectively. Aside from \name{}, we also run speculative decoding (``SpecDecode'') with the smaller model as the draft model, speculating five tokens at a time. To demonstrate \name{}'s compatibility with speculative decoding, we also run a ``\name{}+Decode'' baseline that employs the hierarchical speculation described in~\S\ref{sec:hierarchical_speculation}.

\textbf{Hardware.} We run our evaluations on two NVIDIA A6000-48GB GPUs. We use vLLM~\citet{vllm} 0.8.2 as the underlying inference engine and enable prefix caching. Both models are served with a tensor parallelism degree of two.  %

\begin{figure}[t]
    \centering
    \begin{subfigure}[b]{\textwidth}
        \includegraphics[width=0.99\textwidth]{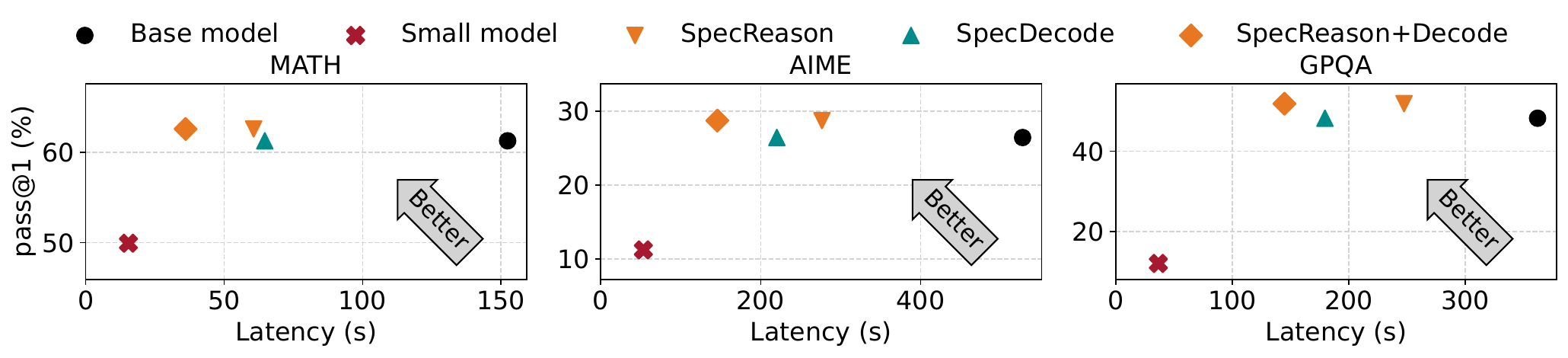}
        \vspace{-5pt}
        \caption{QwQ-32B + R1-1.5B}
        \label{fig:main_results_1}
    \end{subfigure}
    \hfill
    \begin{subfigure}[b]{\textwidth}
        \includegraphics[width=0.99\textwidth]{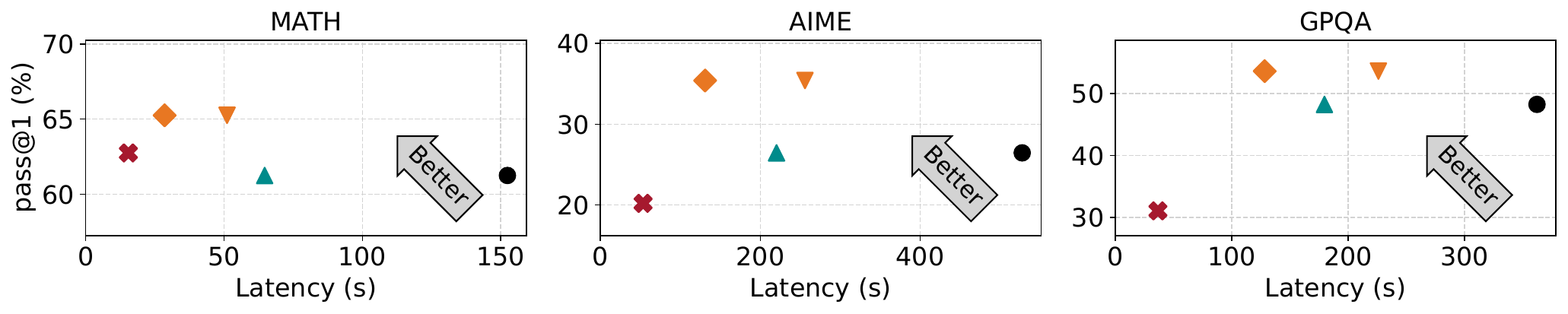}
        \vspace{-5pt}
        \caption{QwQ-32B + Zyphra-1.5B}
        \label{fig:main_results_2}
    \end{subfigure}
    \hfill
    \begin{subfigure}[b]{\textwidth}
        \includegraphics[width=0.99\textwidth]{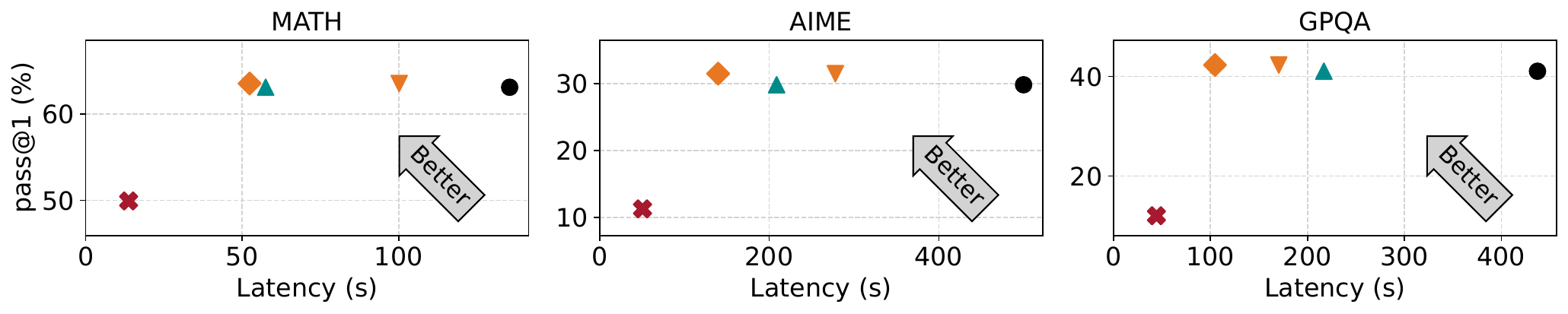}
        \vspace{-5pt}
        \caption{Skywork-Preview-32B + R1-1.5B}
        \label{fig:main_results_3}
    \end{subfigure}
    \hfill
    \begin{subfigure}[b]{\textwidth}
        \includegraphics[width=0.99\textwidth]{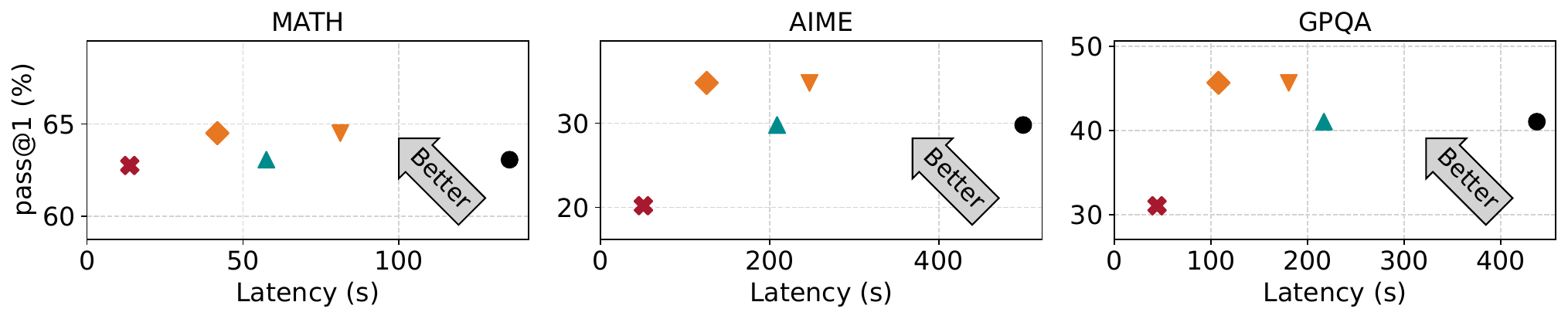}
        \vspace{-5pt}
        \caption{Skywork-Preview-32B + Zyphra-1.5B}
        \vspace{-5pt}
        \label{fig:main_results_4}
    \end{subfigure}
    \caption{Comparison of the accuracy and latency of different schemes on different model combinations. \name{} significantly reduces latency while improving accuracy over vanilla inference. When combined with speculative decoding, \name{} outperforms speculative decoding in both latency and accuracy on all datasets and model combinations.}
    \label{fig:main_results}
    \vspace{-15pt}
\end{figure}

\begin{figure}[t]
    \centering
    \includegraphics[width=0.99\textwidth]{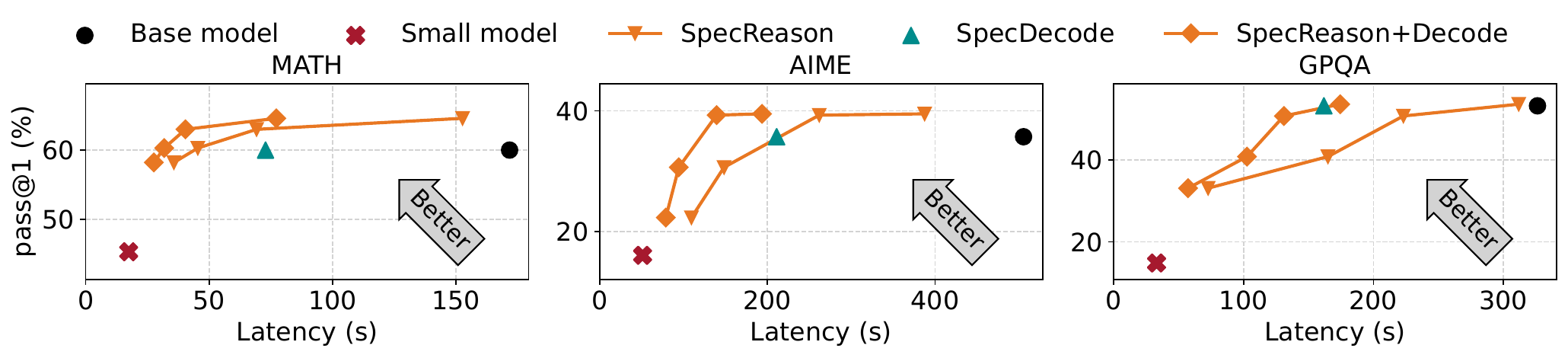}
    \caption{[QwQ-32B + R1-1.5B] \name{} allows trading off latency for accuracy via adjusting the acceptance threshold (from left to right, the thresholds are: 3, 5, 7, and 9 out of 9).}
    \label{fig:lat_acc_tradeoff}
    \vspace{-10pt}
\end{figure}

\subsection{Main Results}
\label{sec:main_results}

We compare \name{} against baseline methods in Fig.~\ref{fig:main_results}. Across the four model combinations, \name{} achieves a 1.5$\times$--2.5$\times$, 1.6$\times$--3.0$\times$, 1.4$\times$--2.5$\times$, 1.7$\times$--2.4$\times$ reduction in latency, respectively, compared to vanilla inference with the base model.   %

\textbf{Accuracy improvement.} Alongside these efficiency gains, \name{} also yields modest accuracy improvements of 1.3\%--3.6\%, 4.0\%--9.0\%, 0.4\%--1.7\%, and 1.4\%--5.0\% compared to the base model. The key reason behind this accuracy improvement is the reduction in token consumption required for reasoning. In Fig.~\ref{fig:acc_intuition}, we focus on the model combination with the highest overall accuracy improvement, QwQ-32B + Zyphra-1.5B, and compare the average number of thinking tokens needed to derive an answer between the base model, the small model, and \name{}. As seen in Fig.~\ref{fig:output_len_comparison}, the small model is generally less verbose than the base model, and because \name{} adopts many speculated steps from the small model, its token consumption is also reduced by 1.2$\times$--2.0$\times$. We also focus on the AIME dataset and vary the token budget to study its effect on the difference in accuracy between \name{} and the base model in Fig.~\ref{fig:acc_by_seq_len}. The effect of token reduction on accuracy is the most significant for tighter output token budgets (16.2\% at 4096 tokens) but shrinks as the base model is allowed to generate more thinking tokens (4.7\% at 8192 tokens). We also attribute these accuracy gains to \name{}’s explicit judgment and scoring mechanism at each reasoning step, which augments the model’s internal self-reflection with more structured assessment.

When compared with speculative decoding, \name{} lies on the Pareto frontier of the accuracy-latency tradeoff. More importantly, combining \name{} with speculative decoding (\name{}+Decode) results in further latency reductions of 19.4\%--44.2\%, 30.8\%--58.0\%, 8.8\%--52.2\%, and 25.1\%--51.8\% over speculative decoding alone.
The most significant performance gains for \name{} when the base model is QwQ-32B occur on the MATH dataset, where both models achieve relatively high accuracies and the capability gap between the small and base models is the narrowest. This makes intermediate steps easier for the small model to speculate correctly, increasing the acceptance rate of speculated steps and thereby lowering end-to-end latency. In comparison, Skywork-Preview-32B is slightly inferior at instruction following, so \name{} has to adopt a higher threshold to avoid an accuracy loss, reducing \name{}'s latency wins.

Finally, when comparing \name{}+Decode with \name{}, \name{}+Decode reduces latency by 1.7$\times$--1.9$\times$, 1.7$\times$--1.8$\times$, 1.6$\times$--2.2$\times$, and 1.6$\times$--2.1$\times$, demonstrating the difference in ease of speculation across varying tasks. On these three datasets, the ratio of steps carried out by small models in \name{} is 38.1\%--80.0\%, 36.5\%--71.3\%, 39.3\%--70.2\%, and 41.4\%--66.6\%, respectively.

\subsection{Controlling the Accuracy-Latency Tradeoff}
\label{sec:acc_lat_tradeoff}

In Fig.~\ref{fig:lat_acc_tradeoff}, we illustrate how \name{} enables flexible control over the accuracy-latency tradeoff, using a representative, randomly selected subdataset from the full datasets in \S\ref{sec:main_results} on QwQ-32B + R1-1.5B for ease of evaluation. During the base model’s evaluation of each reasoning step, we vary the acceptance threshold for the utility score between 3, 5, 7, and 9, and report the resulting accuracy and latency.

On the MATH subdataset, increasing the acceptance threshold from 3 to 7 results in fewer speculative steps from the small model being accepted. This leads to a latency increase from 35.7s to 69.2s, while accuracy improves from 59.4\% to 63.7\%, due to tighter control over the approximation level of intermediate reasoning steps. Notably, the gap between SpecReason+Decode and SpecReason widens from 8.1s to 28.8s, since more reasoning steps are delegated to the base model, and SpecReason+Decode reduces only the base model’s decoding time compared to \name{}.

A similar trend is observed on the AIME and GPQA subdatasets: as the acceptance threshold increases from 3 to 7, latency grows from 109.4s to 261.9s and from 72.7s to 223.0s, and accuracy improves from 22.3\% to 39.3\% and from 33.1\% to 50.7\%. However, the accuracy degrades less gracefully as the threshold is relaxed compared to the MATH subdataset. This is because the small model exhibits a larger performance gap relative to the base model on AIME and GPQA, making aggressive acceptance of its speculative steps more costly in terms of accuracy.

\begin{figure}[t]
    \centering
    \begin{minipage}{0.42\textwidth} %
        \centering
        \includegraphics[width=\textwidth]{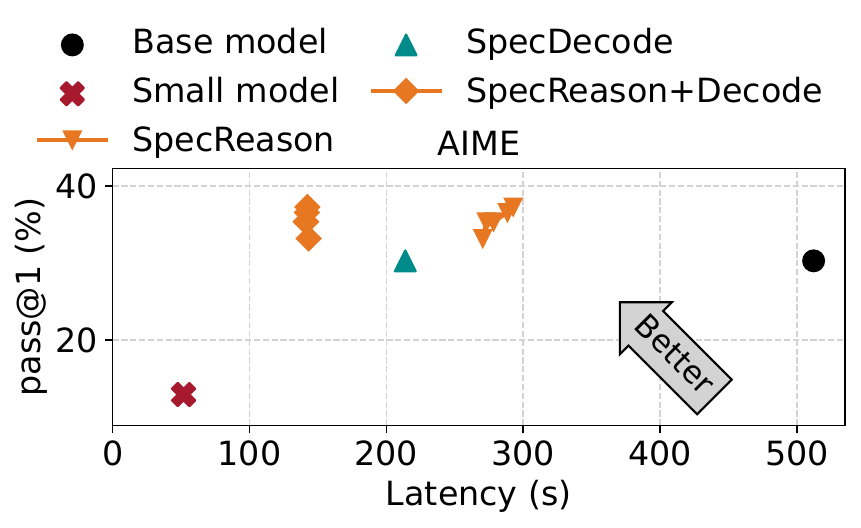}
        \caption{Effect of the alternative knob: forcing the first $n$ steps for base model decoding.}
        \label{fig:earlyforce_knob}
        \vspace{-10pt}
    \end{minipage}
    \hfill %
    \begin{minipage}{0.55\textwidth} %
        \centering
        \includegraphics[width=\textwidth]{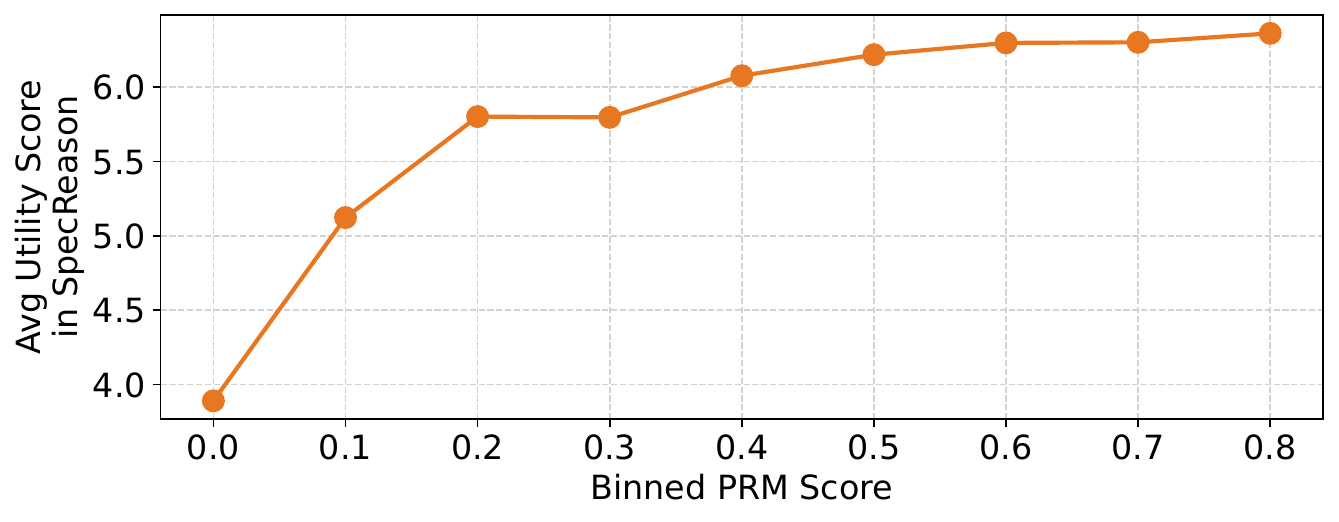}
        \caption{The utility scores in \name{} closely reflect the quality score judgements from a process reward model. $x$ on the x-axis denotes PRM scores in the range $[x, x+0.1)$.}
        \label{fig:prm_microbenchmark}
    \end{minipage}
\end{figure}

In Fig.~\ref{fig:earlyforce_knob}, we also study the effect of the alternative knob, forcing the first $n$ reasoning steps to be decoded by the base model, on the accuracy-latency tradeoff. As we change $n$ from 0 to 10, 20, 30, and 40, \name{}'s accuracy increases from 33.2\% to 37.3\% while the latency increases from 270.4s to 292.6s, showcasing an alternative approach to improve accuracy with a slight increase in latency.

\subsection{Base Model's Judgement Capability}

The base model's ability to assess the quality of intermediate reasoning steps is a crucial cornerstone of \name{}'s performance. In this experiment, we compare the scores generated by a process reward model (PRM) -- which assigns a reward score to each step within the solution to a math problem -- with those given by the QwQ-32B base model on the AIME dataset. Specifically, we use Math-Shepherd~\citep{math_shepherd}, a PRM trained via reinforcement learning from the Mistral-7B base model on math problems, to score each speculated step produced by the R1-1.5B small model.

In Fig.~\ref{fig:prm_microbenchmark}, we bin the reward scores (a float from 0 to 1) into ten bins. Within each bin, we calculate the mean utility score given by the base model in \name{}. This analysis demonstrates a strong correlation between the base model's and the PRM’s assessments, particularly for lower-quality reasoning steps, where both models assign low scores. The results suggest that the base model can effectively approximate the PRM’s judgments, making it a viable option for evaluating reasoning step quality in \name{}.

\section{Conclusion}

In this work, we introduce \name{}, a novel approach that accelerates LRM inference by leveraging speculative reasoning. By offloading simpler intermediate reasoning steps to a smaller, lightweight model and reserving the base model for assessment, \name{} significantly reduces inference latency while maintaining or even improving accuracy. Our results demonstrate that \name{} achieves a $1.4-3.0\times$ speedup over vanilla LRM inference, with accuracy improvements ranging from $0.4-9.0\%$. Additionally, when combined with speculative decoding, \name{} further reduces latency by $8.8-58.0\%$, highlighting the complementary nature of these optimizations. We believe this work opens up new angles for efficient LRM inference acceleration, making it especially valuable for scenarios that demand both high accuracy and low latency.

\begin{ack}
We thank Princeton's Systems for Artificial Intelligence Lab (SAIL) and Princeton Language and Intelligence (PLI) for providing the hardware resources for running experiments. This work was supported by NSF CNS grants 2147909, 2151630,
2140552, 2153449, and 2152313.
\end{ack}

{
\small
\bibliography{references}
\bibliographystyle{plainnat}
}

\appendix

\newpage{}
\section{Appendix}
\label{sec:appendix}

\subsection{Base Models of Varying Sizes and Architectures}
\label{sec:70b_microbenchmark}

\begin{figure}[h]
    \centering
    \includegraphics[width=0.6\textwidth]{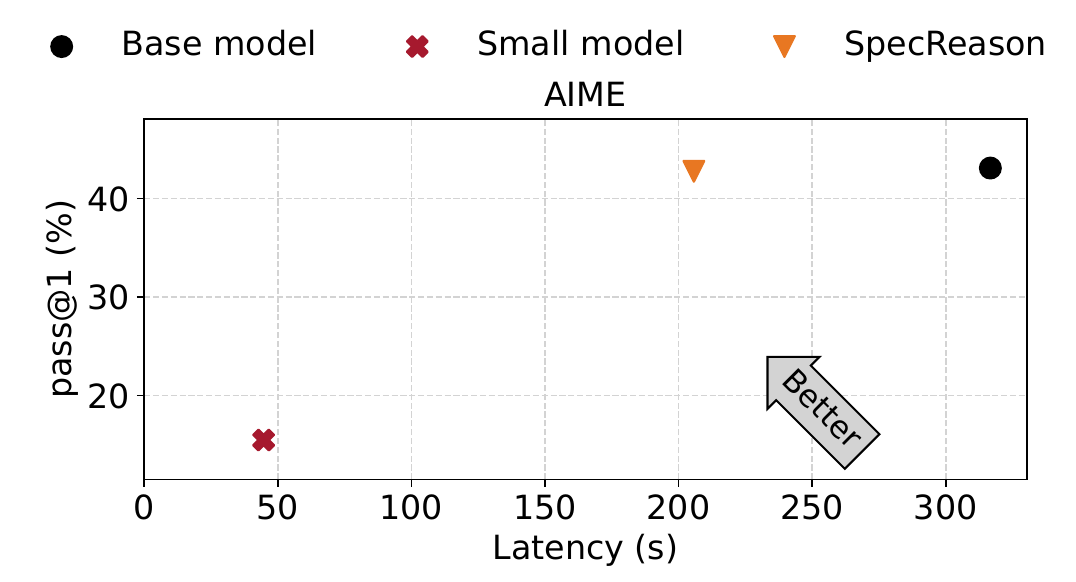}
    \caption{\name{}'s results on the model combination (R1-70B, R1-1.5B).}
    \label{fig:70b_results}
\end{figure}

To demonstrate the generality of \name{}, we replace the QwQ-32B base model with DeepSeek's R1-70B and evaluate on the same representative subdatasets as in~\S\ref{sec:acc_lat_tradeoff}. Given the size of the R1-70B model, we deploy it across four A100-80GB GPUs using a tensor parallelism degree of 4.

On the AIME subdataset, \name{} achieves a 1.5$\times$ latency reduction compared to vanilla R1-70B inference. This speedup is smaller than the gains observed with the QwQ-32B model in our main results (1.9$\times$) due to two key factors. First, the R1-70B model benefits from both stronger hardware and greater parallelism (4-way TP on A100s), resulting in a 1.5$\times$ lower time-per-token (TPT) compared to QwQ-32B (2-way TP on A6000s). In contrast, the smaller model R1-1.5B sees only a modest 1.1$\times$ TPT improvement on stronger hardware, which narrows the performance gap between base and small models and thus diminishes latency savings. Second, QwQ-32B is empirically a stronger model -- outperforming R1-70B across many reasoning benchmarks~\cite{qwq_32b} -- and this performance gap impacts their respective abilities to assess intermediate steps. To maintain accuracy, we adopt a stricter acceptance threshold when using R1-70B as the base model, which reduces the fraction of steps offloaded to the small model (23.2\% compared to 40.8\% in the main results).

\subsection{Intuition behind Accuracy Improvement}

\begin{figure}[h]
    \centering
    \includegraphics[width=0.99\textwidth]{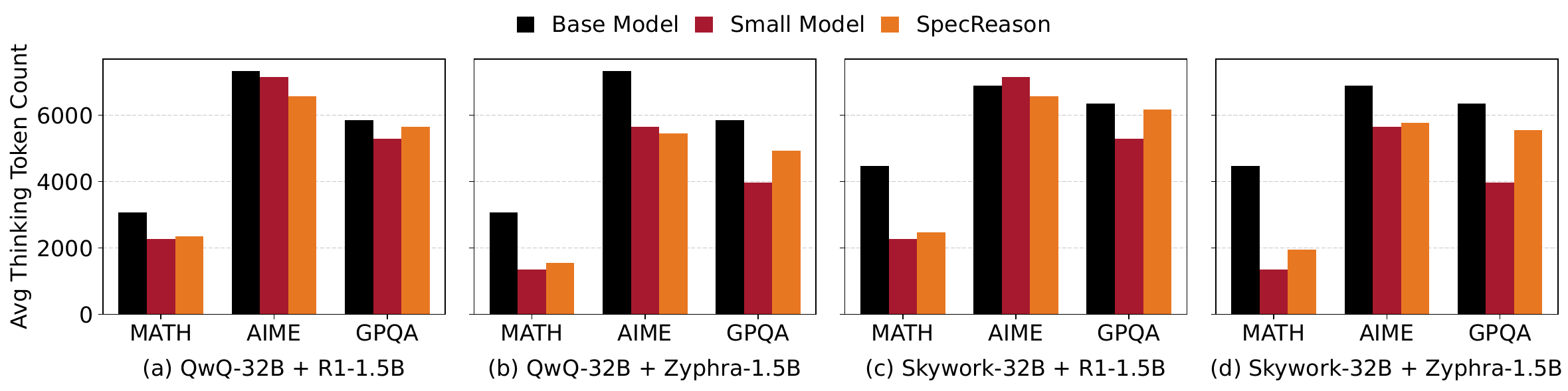}
    \caption{Intuition behind \name{}'s accuracy improvement on all datasets and model combinations.}
    \label{fig:acc_insight_len_all}
\end{figure}

In Fig.~\ref{fig:acc_insight_len_all}, we evaluate the average thinking token count of \name{} and two vanilla inference baselines on a wide range of datasets and model combinations. We observe that the small model is generally less verbose than the base model, and because \name{} adopts many speculated steps from the small model, its token consumption is reduced by $1.0-1.3\times$, $1.2-2.0\times$, $1.0-1.8\times$, and $1.1-2.3\times$ on the four model combinations, respectively.

\end{document}